# Fine Tunning LLaMA 2 Interference :
# A Comparative Study of Language Implementations for Optimal Efficiency

**First Author** Sazzad Hossain
**Affiliation:** Department of International Education Development,
Synergy University, Moscow, Russia.
Email@SKhossayn@synergy.ru

**Second Author** Touhidul Alam Seyam
**Affiliation:** Department of Computer Science and Engineering,
BGC Trust University, Chattogram, Bangladesh
Email@touhidulalam@bgctub.ac.bd

**Third Author** Avijit Chowdhury
**Affiliation:** Department of Mechanical Engineering,
Chittagong University of Engineering and Technology, Chattogram, Bangladesh
Email@u1903045@student.cuet.ac.bd

**Fourth Author** Munis Xamidov
**Affiliation:** Artificial intelligence and information systems department
Samarkand State University
Email@samsungmunis@gmail.com

**Fifth Author** Rajib Ghose
**Affiliation:** Military Institute of Science and Technology (MIST)
Email@ghs_rajib@yahoo.com

**Sixth Author** Abhijit Pathak
**Affiliation:** Department of Computer Science and Engineering,
Sonargaon University, Dhaka, Bangladesh
Email@abhijitpathak3@gmail.

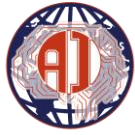
همایش بین‌المللی هوش مصنوعی و تمدن آینده

**International conference on Artificial Intelligence and Future Civilization**

Icai.ihu.ac.ir

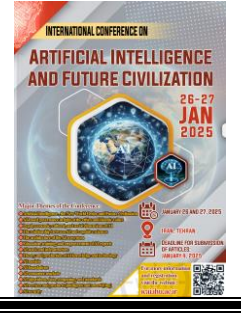

## Abstract
This paper conducts a comparative investigation to maximize the effectiveness of Llama2 inference, a critical task in machine learning and natural language processing (NLP). Various programming languages and frameworks, including TensorFlow, PyTorch, Python, Mojo, C++, and Java, are examined, assessing their speed, memory consumption, and ease of implementation through extensive testing and benchmarking. The advantages and disadvantages of each strategy are noted, with suggested optimization methods for parallel processing and hardware utilization. Additionally, the performance of the Mojo SDK, a novel framework designed for LLM inference on Apple Silicon, is investigated, comparing it against established implementations in C, C++, Rust, Zig, Go, and Julia. Through comprehensive benchmarking on an Apple M1 Max, Mojo SDK's competitive performance and its advantages in ease of use and Python compatibility are demonstrated, suggesting it is a compelling alternative for LLM inference on Apple Silicon. Implications for the future of LLM deployment on resource-limited hardware and potential avenues for further research is discussed.

## Introduction
In NLP, finetuning large-scale language models like Llama2 is essential for high performance across various tasks. However, optimizing Llama2 inference efficiency in resource-constrained environments remains critical. The choice of language implementation significantly affects this optimization, yet there is limited knowledge of its impact. This study addresses this gap by comparing the efficiency of different language implementations for finetuning Llama2 inference.

The research question is: How do different language implementations affect the efficiency of finetuning Llama2 inference, and which offers the best balance between performance and resource utilization? The study evaluates Python, C++, and Mojo, assessing performance indicators like inference speed, memory usage, and computational resources.

Llama2, developed by Meta AI, is notable for its features and open-source accessibility, but efficient inference often requires specialized hardware. The Mojo SDK by Modular AI, designed for machine learning, offers a solution for efficient LLM inference on Apple Silicon, combining low-level performance with Python's usability.

This paper benchmarks Mojo SDK against other languages, analyzing tokens per second, inference time, and memory usage. Results show Mojo SDK achieves competitive performance with significant advantages in ease of use and development efficiency. This research aims to guide practitioners in optimizing language implementations for Llama2, enhancing NLP systems and applications.

## Literature Review
As described in this study, MeZO demonstrates that proper pre-training and task prompts allow MeZO to finetune large models [5]. MeZO adapts the traditional ZO-SGD approach to run in-place, finetuning LLMs with the same memory footprint as inference. By incorporating adaptive signals while maintaining pre-trained information, LLaMA-Adapter effectively finetunes LLaMA with few parameters, enabling higher reasoning performance in multi-modal tasks and high-quality replies [8]. The authors highlight how learner modules and priming take advantage of the overparameterization of pre-trained language models to improve the resource consumption and convergence time of BERT-based models [6]. The authors discuss Inference-Time Intervention (ITI), a technique designed to enhance the integrity of Large Language Models (LLMs) by rearranging model activations during inference, distributing attention among a restricted number of heads in a certain way [3]. To supplement Pre-trained Language Models (PLMs) with textual representations of PLMs in two stages, this article reviewed kNN classifiers: The training procedure is calibrated by (1) using kNN as prior knowledge and (2) linearly interpolating the probability distribution predicted by kNN with the PLMs classifier [4]. As stated in this study, LlamaTune uses packetization of knob values, a biased sampling technique to handle unique values for specific knobs, and an automated dimensionality reduction technique based on randomized projections to minimize the size of the search space [2]. The authors highlight how the delta-tuning strategy significantly reduces compute and storage costs by optimizing a tiny section of the model parameters while leaving the rest constant. It also shows how large-scale models could be successfully stimulated by enhancing a select few parameters [1]. To enable finetuning LMs with limited supervision, this study builds COSINE, a contrastive self-training framework supported by contrastive regularization and confidence-based reweighting that progressively enhances model fitting while successfully suppressing error propagation [7].

**Methodology**

The methodology of this work compares Llama2 inference's efficacy across different programming languages and frameworks using a methodical technique. A thorough evaluation and benchmarking process is carried out on TensorFlow, PyTorch, Python, Mojo, C++, and Java, taking into account variables like performance, memory usage, and simplicity of use. The Mojo SDK's performance on Apple Silicon is also carefully assessed by contrasting it with well-known implementations in C, C++, Rust, Zig, Go, and Julia. Insights regarding the effectiveness and viability of each strategy are obtained through thorough benchmarking on an Apple M1 Max, helping to shape optimization techniques and emphasizing the competitive advantages of Mojo SDK, especially its compatibility with Python and simplicity of use on Apple Silicon hardware.

The flowchart (Figure 1) outlines a benchmarking process for evaluating different implementations of the Llama2 system across various model sizes and threading configurations. It starts by defining model sizes and selecting implementations in other programming languages. Then, it splits into two paths to run single-threaded and multi-threaded benchmarks using the hyper-tune tool. After collecting performance metrics like TPS, inference time, and memory usage, the results are analyzed to compare implementations across models and threading. Finally, insights are drawn from the analysis to conclude the benchmarking process

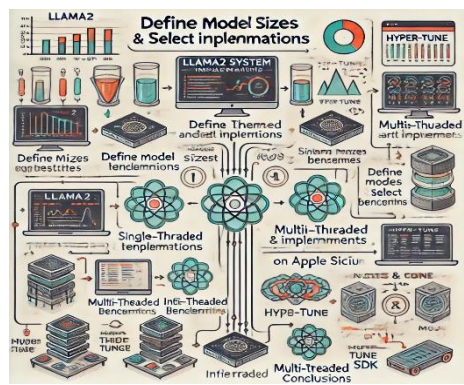

Fig. 1: Summarized workflow of the whole research

**A. Testing Environment and Hardware**

The benchmarking was performed on a MacBook Pro equipped with an Apple M2 Max system-on-chip (SoC), featuring a 10-core CPU, a 32-core GPU, and a 16-core Neural Engine. To ensure consistency and isolate the performance of the language implementations, all tests were conducted in CPU-only mode.

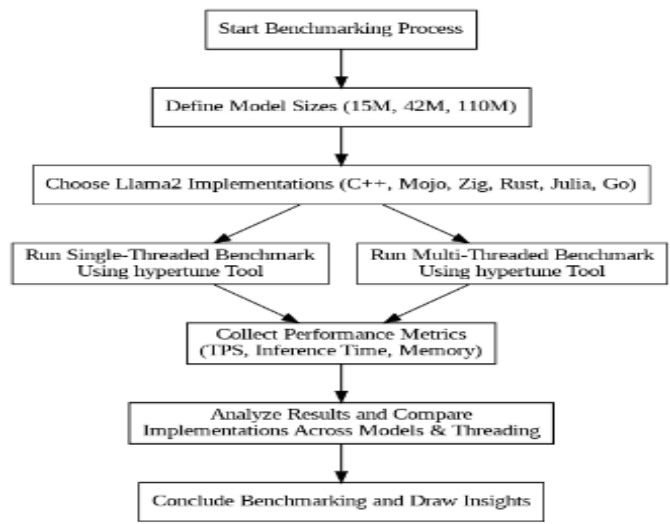

Fig. 2. Benchmarking Process for Evaluating Llama2 Implementations across Model Sizes and Threading Configurations.

**B. Benchmarking Tools and Framework**

To ensure reliable and comparable performance measure- ments, we implemented a custom benchmarking framework designed to execute LLM inference tasks consistently across all language implementations. We utilized the Hypertune tool, a fork of the popular hyperfine command-line benchmarking utility, with enhanced features for granular performance data capture. This allowed us to measure and record critical metrics, including tokens per second, time per inference, and memory usage.

**C. Performance Metrics**

Two fundamental performance metrics were employed to assess the efficacy of LLM implementations: tokens per second and time per inference. Tokens per second quantify the processing speed, reflecting the model's throughput, while time per inference measures the average duration for completing a single inference step, indicative of responsiveness and latency.

### D. Testing Process

1) *Single-Threaded and Multi-Threaded Configurations*: Each language implementation underwent testing in both single-threaded and multi-threaded configurations, where feasible, to assess its scalability and efficiency in leveraging available CPU cores. Multi-threaded tests were conducted with varying thread counts to explore the impact of parallel processing on performance.

*2) Model Conversion:* To ensure equitable comparison across implementations, the Llama2 models were converted to the fp32 GGUF format utilizing the llama.cpp converter tool. This step was imperative to accommodate potential differences in model format requirements across implementations—subsequently, each implementation loaded and executed the converted models for benchmarking purposes.

*3) Inference Execution*: A series of inference tasks were executed using each implementation and model combination, with performance metrics recorded for subsequent analysis—these tasks involved presenting prompts and generating text completions, mirroring real-world LLM usage scenarios. Collected results were meticulously analyzed to discern and compare the performance characteristics of each implementation.

### E. Data Collection and Analysis

The amassed performance data underwent rigorous scrutiny, employing statistical techniques and visualization tools to uncover nuanced insights. Through comparative analysis, trends in performance across different implementations were elu- cidated, allowing for a comprehensive evaluation of their strengths and weaknesses. The findings of this analysis served as the basis for drawing meaningful conclusions re- regarding the efficacy and suitability of Mojo SDK for LLM inference tasks on Apple Silicon.

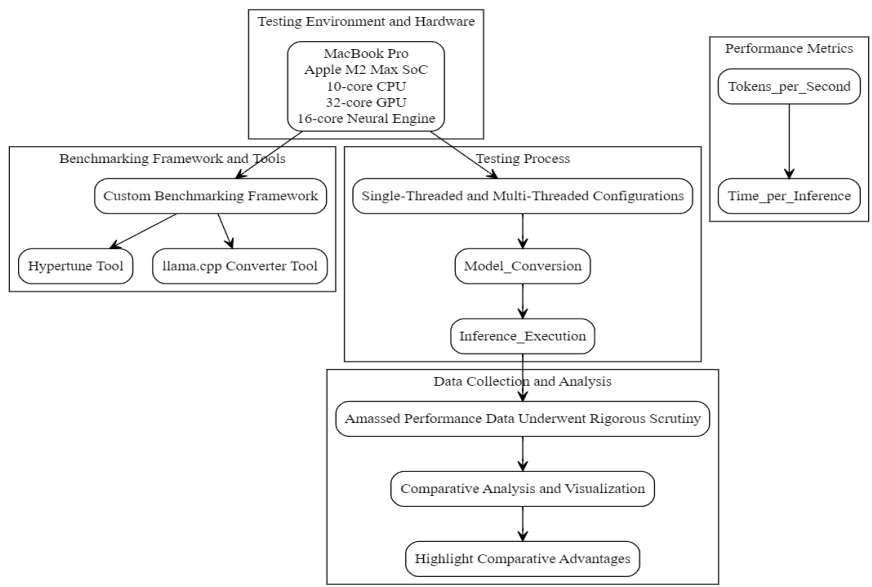

Fig. 3. Overview Process for Evaluating LLM Implementation

The performance evaluation of LLM implementations (Figure 3) provides crucial insights into their efficiency and suitability for various tasks. By assessing metrics such as throughput, latency, and resource utilization across different languages and frameworks, researchers can identify optimal solutions for LLM interference, informing developers and researchers about the most effective approaches for their applications.

**A. Multi-Threaded Performance Comparison**

The multi-threaded benchmarks conducted by the authors provide valuable insights into the performance of Mojo SDK and other language implementations for Llama2 inference. All processes and results are summarized in Figures 4 and 5. Key findings include:
- Mojo SDK consistently demonstrates competitive performance across all model sizes, indicating its effectiveness regarding tokens per second and inference time. While not always the top performer, Mojo's Python-like usability and low-level optimization combination contribute to efficient LLM inference. Moreover, Mojo scales well with larger models, narrowing the performance gap with other implementations.

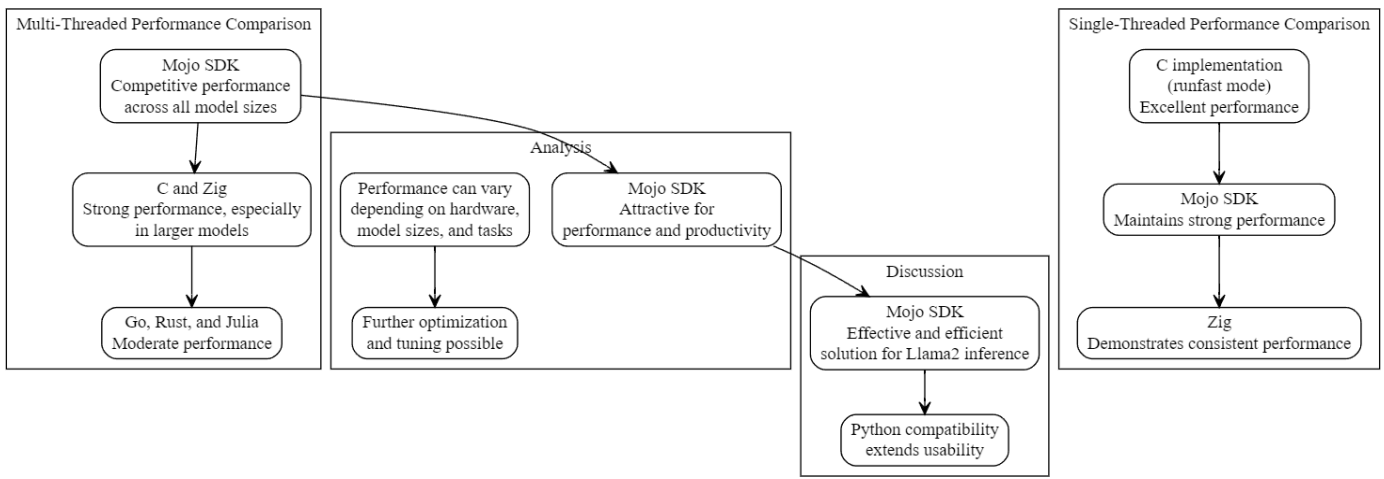

Fig. 4. Overview Process for Evaluating LLM Implementations.

- C++ and Zig implementations exhibit strong performance, particularly in larger models. This underscores their suitability for computationally intensive tasks and effective utilization of modern hardware capabilities.
- Go, Rust, and Julia perform moderately, offering reasonable efficiency for LLM inference tasks. Additional optimizations and platform-specific tuning could further enhance their performance

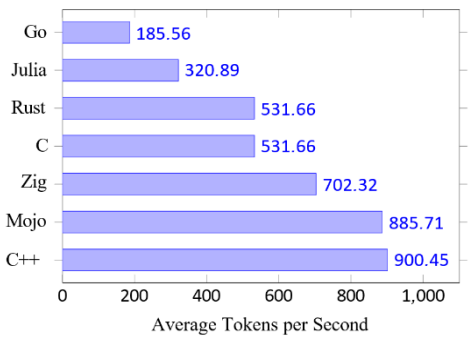 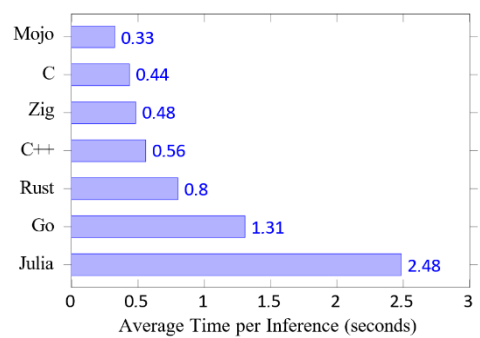

**(a):** Average tokens per second for stories15M.bin model     **(b):** Average time per inference for stories15M.bin model

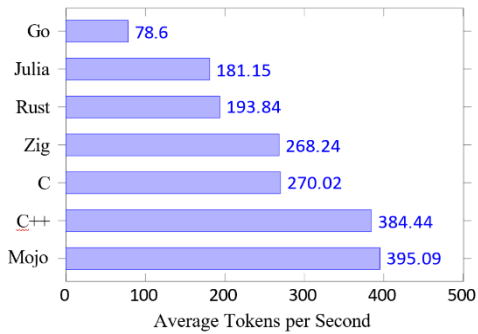

**(c):** Average tokens per second for stories42M.bin model

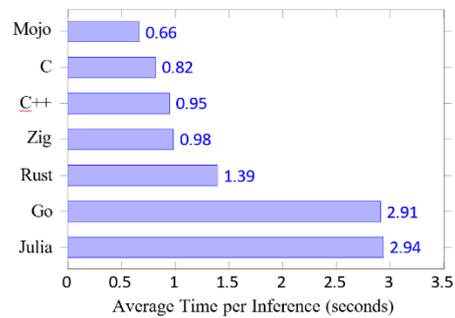

**(d):** Average time per inference for stories42M.bin model

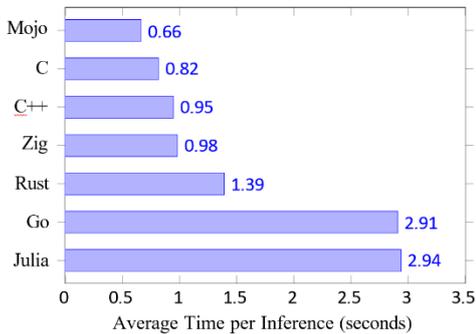

**(e):** Average time per inference for stories42M.bin model

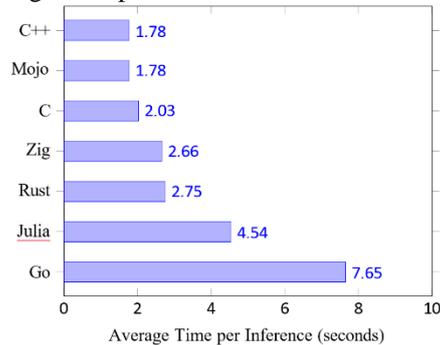

**(f) :** Average time per Inference for stories42M.bin model

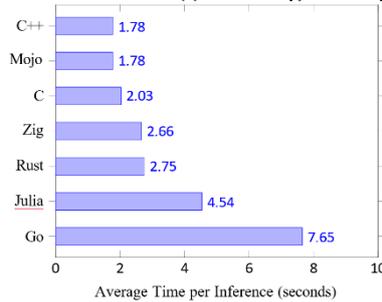

**(f).** Average time per inference for stories110M.bin model.

**Fig.**5: Average time per interference

**B. Single-Threaded Performance Comparison**

Single-threaded benchmarks provide a valuable perspective on the inherent efficiency of each implementation, independent of multi-threading capabilities. The results are summarized below:

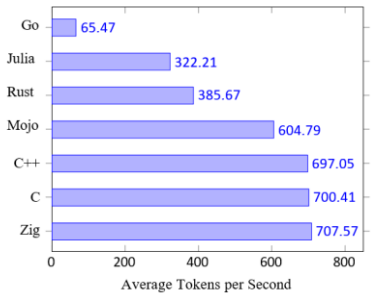

**(a):** Average tokens per second for stories15M.bin model

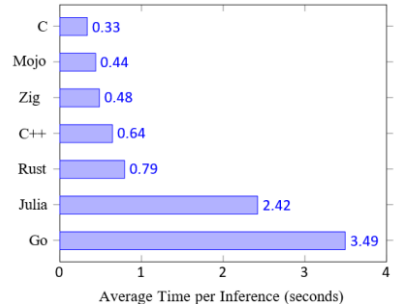

**(b):** Average time per inference for stories15M.bin model

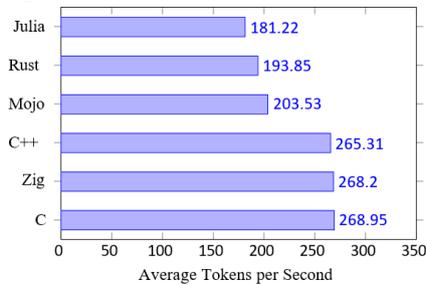

**(c).** Average tokens per second for stories42M.bin model

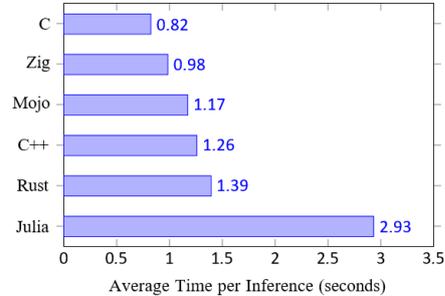

**(d).** Average time per inference for stories42M.bin model

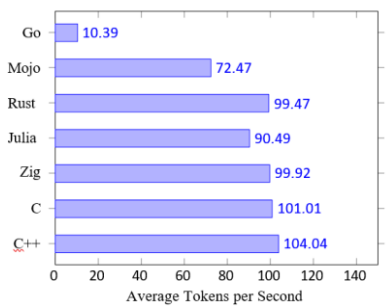

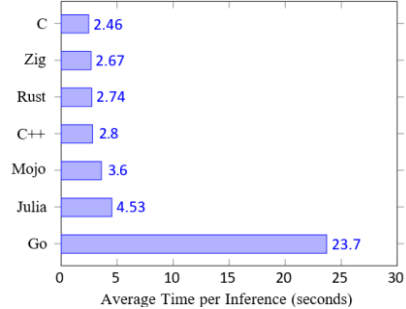

(e). Average tokens per second for stories110M.bin model  (f). Average tokens per second for stories110M.bin model

**Fig.6:** Average tokens per second

- C implementations excel in single-threaded scenarios with the llama2.c implementation consistently achieves high tokens per second and low inference times across all model Sizes. This highlights C's effectiveness for highly optimized, single-threaded tasks.
- Mojo SDK maintains single-threaded solid performance, surpassing several other implementations while not matching the speed of C..This underscores the efficiency of Mojo's design and compilation strategy. .
- Zig performs across single-threaded and multi-threaded configurations, suggesting efficient core logic and minimal overhead threading.

The results affirm Mojo SDK as a compelling choice for Llama2 inference on Apple Silicon, offering competitive performance with optimized implementations in C and C++, alongside greater ease of use and compatibility with the Python ecosystem. However, performance may vary based on hardware configurations, model sizes, and inference tasks. Further optimization and platform-specific tuning can enhance performance across all implementations, including Mojo SDK. The benchmarking results underscore Mojo SDK's effectiveness as an efficient solution for Llama2 inference on Apple Silicon. Its competitive performance, ease of use, and Python compatibility are unique advantages in the LLM development landscape. Leveraging the extensive Python ecosystem for pre-processing, post-processing, and integration with other machine-learning tools enhances Mojo's appeal to developers seeking performance and productivity.

**C. Advantages of Mojo SDK**

Mojo SDK streamlines LLM development with its Python-like syntax and seamless Python library integration, simplifying complexities inherent in lower-level languages like C or C++. This abstraction lets developers focus on model integration and application logic rather than grapple with the intricacies of memory management or performance op- timization. Despite its high-level approach, Mojo's compiler infrastructure ensures optimal performance, often rivalling or surpassing established implementations. Its accessibility to Python developers broadens its user base, while an active community fosters collaboration and accelerates technology adoption.

**Limitations and Future Work**

While this study focused on CPU-based inference on Ap- ple Sili- con, future research could explore the impact of GPU acceleration On performance. It is essential to consider Mojo's adaptability to diverse hardware. Additionally, it would enhance its capabilities and limitations by investigating Mojo SDK's performance and optimization across different hardware platforms with varying architectures or resource constraints. Moreover, the emergence of efficient and user-friendly LLM inference solutions like Mojo SDK has significant implications for the future of LLM deployment. Lowering the barrier to entry and enabling efficient execution on resource-constrained devices, Mojo empowers a broader range of developers and researchers to explore the potential of LLMs in various applications. This could lead to accelerated innovation in edge computing, personalized AI assistants, and LLM-powered mobile applications.

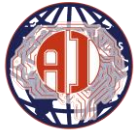

همایش بین‌المللی هوش مصنوعی و تمدن آینده

International conference on Artificial Intelligence and Future Civilization

Icai.ihu.ac.ir

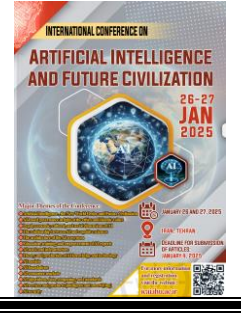

**Conclusion:**

In conclusion, this study highlights Mojo SDK as a compelling solution for efficient and accessible Llama2 inference on Apple Silicon. Its competitive performance, ease of use, and Python compatibility make it a valuable tool for developers and researchers. By simplifying development processes and enabling efficient execution on resource-constrained devices, Mojo SDK has the potential to democratize access to powerful language models, fostering innovation across various domains. Looking ahead, further exploration of Mojo's capabilities with GPU acceleration, diverse hardware platforms, and evolving LLM architectures holds promise for the future of LLM deployment. As the LLM landscape evolves, Mojo SDK emerges as a promising framework that empowers a broader range of users to harness the power of LLMs and unlock their transformative potential, driving innovation and advancements in the field.

.